\documentclass[11pt,a4paper]{article}
\usepackage[hyperref]{emnlp2018}
\usepackage{times}
\usepackage{latexsym}
\usepackage{graphicx}
\usepackage{algorithm}
\usepackage{algpseudocode}
\usepackage{multirow}
\usepackage{amsmath}
\usepackage{tabularx}
\usepackage[skip=1pt]{caption}
\usepackage{url}

\aclfinalcopy 

\usepackage{xcolor}

\title{WikiConv: A Corpus of the Complete Conversational History of a Large Online Collaborative Community}

\author{Yiqing Hua \\
        Cornell Tech \\
       {\tt yiqing@cs.cornell.edu}\\ 
       \AND Cristian Danescu-Niculescu-Mizil \\
       Cornell University \\
       {\tt cristian@cs.cornell.edu}\\\And
       Dario Taraborelli \\
       Wikimedia Foundation \\
       {\tt dtaraborelli@wikimedia.org}
        \\\AND
        Nithum Thain \and Jeffery Sorensen \and Lucas Dixon \\
        Jigsaw, Google \\
        {\tt {nthain, sorenj, ldixon}@google.com}}

\date{}

\begin{document}
\maketitle
\begin{abstract}
We present a corpus that encompasses the complete history of conversations between contributors to Wikipedia, one of the largest online collaborative communities.  
By recording the intermediate states of conversations---including not only comments and replies, but also their modifications, deletions and restorations---this data offers an unprecedented view of online conversation.
This level of detail supports new research questions pertaining to 
the process (and challenges) of large-scale online collaboration.   
We illustrate the corpus' potential with two case studies that highlight new perspectives on earlier work. 
First, we explore how a person's conversational behavior depends on how they relate to the discussion's venue. 
Second, we show that community moderation of toxic behavior happens at a higher rate than previously estimated.
Finally the reconstruction framework is designed to be language agnostic, and we show that it can extract high quality conversational data in both Chinese and English.
\end{abstract}
\section{Introduction}
Compared to large-scale collections of conversations from social media \cite{felbo2017using, luo2012opinion,zhang2017characterizing,tan2016winning} 
or news comments \cite{napoles2017finding},
Wikipedia talk pages offer a unique perspective into goal-oriented discussions between thousands of volunteer contributors coordinating to write the largest online encyclopedia. 
Talk page data already underpins research on social phenomena such as conversational behavior \cite{danescu2012echoes,danescu2013computational}, disputes \cite{wang2014piece}, antisocial behavior \cite{wulczyn2017ex,zhang2018conversations} and  collaboration \cite{kittur2007he, halfaker2009jury}. 
However, the scope of such studies has so far been limited by a view of the conversation that is incomplete in two crucial ways: first, it only captures a subset of all discussions;
and second, it only accounts for the final form of each conversation, which frequently differs from the interlocutors experience as the conversation develops.

In this paper, we undertake the challenge of reconstructing a complete and structured history of the conversational process in Wikipedia talk pages, containing detailed information about all the interlocutors' actions, such as adding and replying to comments, modifying or deleting them. 
To this end, we devise a methodology for identifying and structuring these actions, while also addressing the challenges spurring from the inconsistent formatting and the raw scale of existing records.
This results in the largest public dataset of goal-oriented conversations,
{\em{WikiConv}}, spanning five languages.
The largest component of this dataset is based on the English Wikipedia, and contains roughly 91 million conversations consisting of  212 million conversational actions taking place in 24 million talk pages.  

By including details about how each conversation evolved,
this corpus provides an unprecedented view into the conversational process, as experienced by the interlocutors.
In fact, we find that about $40\%$ of discussion activity would be missed by approaches that do not consider comment modifications and deletions, and even more is missed when only considering the (final) static snapshots of conversations.
Furthermore, a manual review of the English Wikipedia portion of the dataset reveals that $98\%$ of the reply structure is recovered correctly and $98\%$ of the interlocutor's actions are categorized correctly. 

Since the reconstruction pipeline does not rely on any language specific heuristics, we also apply it to Chinese, German, Greek and Russian Wikipedia Talk page archives, in addition to those from English Wikipadia. 
A manual review of the conversations obtained from the Chinese Wikipedia Talk pages shows a similarly high reconstruction accuracy with that obtained from the English Wikipedia, suggesting that it is reasonable to apply the reconstruction pipeline to different languages.
To encourage further validation, refinements and updates, we have open sourced the code and published the datasets.\footnote{\url{https://github.com/conversationai/wikidetox/tree/master/wikiconv}}

Finally, we present two case studies illustrating how the corpus can bring new insights into previously observed phenomena.
We first analyze the conversational behavior of a subset of English Wikipedia contributors across the entire range of talk pages, and show that their levels of linguistic coordination vary according to where the conversation takes place.
Second, we investigate the toxicity of deleted comments, and show that community moderation of undesired behavior takes place at a much higher rate than previously estimated.
\section{Further Related Work}

\begin{figure*}[ht!]
\includegraphics[width=\textwidth]{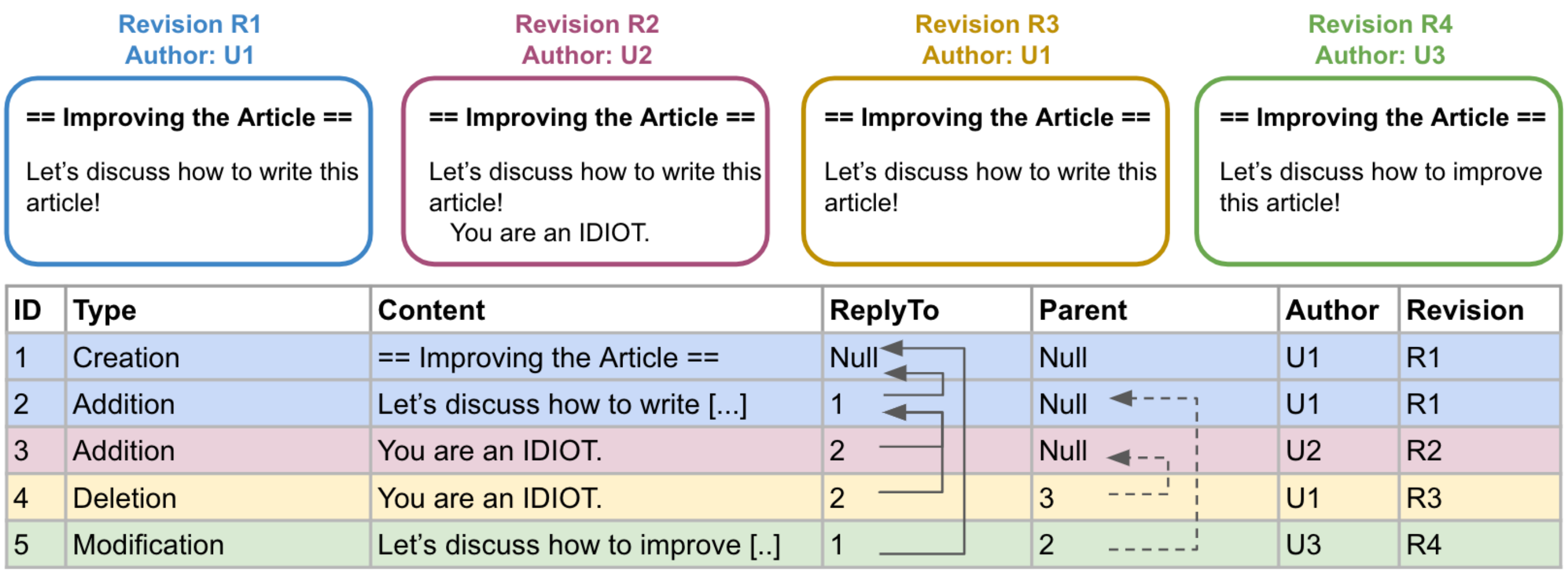}
\caption{\label{toy-example}
\textbf{Example conversation reconstruction.}
The action id in the ReplyTo column defines the conversation's structure; The Parent column indicates history, showing how actions change earlier actions.
Note that each revision (color-coded) can introduce multiple actions.}
\end{figure*}

Past efforts aimed at characterizing conversations on Wikipedia talk pages have either focused on snapshots of discussion threads \cite{danescu2012echoes, prabhakaran2016corpus,wang2014piece,wang2014improving}, or have considered text segments in talk page history as incremental comments, ignoring conversational turns and reply structures within these conversations \cite{wulczyn2017ex}.
The limitations of these approaches can be seen in Figure~\ref{toy-example}, where, if we limit our analysis to only a snapshot of the final state of the conversation, we miss the abusive comment introduced in revision 3 and removed in revision 4, and thus miss an important part of the experience of the participants.  In fact, this ``hidden'' activity accounts for one third of all actions taken on talk pages in English Wikipedia.

The closest dataset to our work is~\citet{bender2011annotating} which introduces the Authority and Alignment in Wikipedia discussions corpus (AAWD), containing 365 talk page discussions.
While acknowledging the complexity of conversational behaviors on Wikipedia talk pages, the AAWD work falls short of providing data on the deletions and follow-up changes to existing comments.
Beyond addressing this shortcoming, the dataset we introduce in this paper is many orders of magnitude larger, containing 91 million conversations in English Wikipedia alone.
\begin{figure}[ht!]
\includegraphics[width=0.5\textwidth]{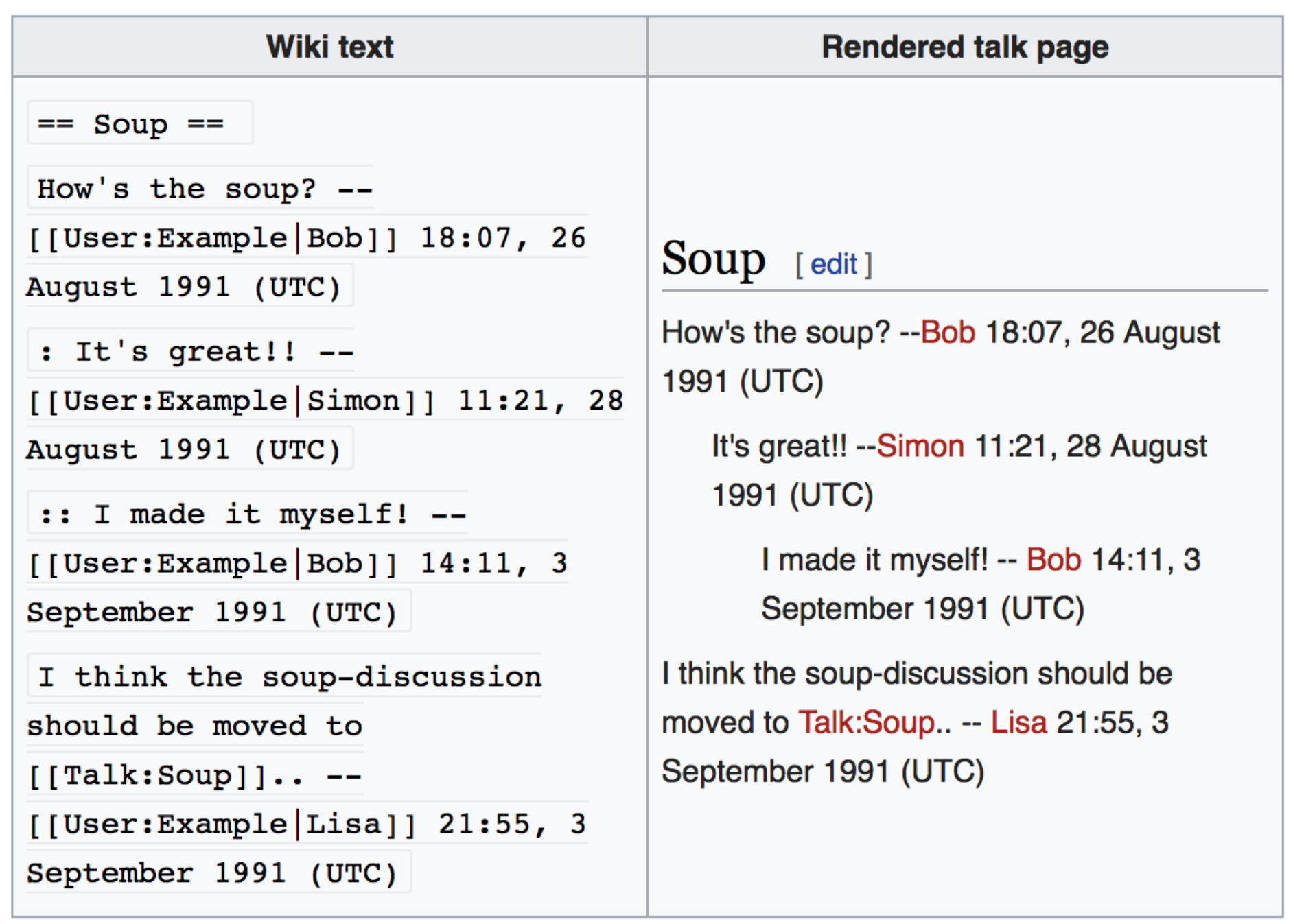}
\caption{\label{talk-example}
An example of Wiki markup and its rendered form from Wikipedia Talk Page Help.\footnotemark}
\end{figure}
\footnotetext{\url{mediawiki.org/wiki/Help:Talk_pages}}

\begin{table*}[ht!]
\centering
\begin{tabular}{|lr|lr|l|l|l|l|}
\hline \multicolumn{4}{|c|}{\it English Wikipedia} & \multicolumn{4}{c|}{\bf Reconstruction Accuracy by Action Type}\\
\hline \multicolumn{2}{|l|}{\bf Number of} &  \multicolumn{2}{l|}{\bf Action Type Breakdown} & \it Boundary & \it Type & \it ReplyTo & \it Parent \\
\hline
Distinct users & $4.4$M & Creation      & $21\%$&$100\%$&$100\%$&$100\%$ &$100\%$\\
Talk Pages     & $24$M  & Addition      & $39\%$&$96\%$&$100\%$&$95\%$ &$100\%$\\
Revisions      & $120$M  & Modification  & $13\%$&$97\%$&$95\%$&$97\%$ &$95\%$\\
Conversations  & $91$M  & Deletion      & $24\%$&$94\%$&$96\%$&$100\%$&$100\%$\\
Actions        & $212$M & Restoration   & $3\%$ &$84\%$&$98\%$&$100\%$&$99\%$\\
\cline{3-8}
 & & \multicolumn{2}{r|}{\bf All actions:}      &$96\%$&$98\%$&$98\%$&$99\%$\\ 
\hline \multicolumn{4}{|c|}{\it Chinese Wikipedia} & \multicolumn{4}{c|}{\bf Reconstruction Accuracy by Action Type}\\
\hline \multicolumn{2}{|l|}{\bf Number of} &  \multicolumn{2}{l|}{\bf Action Type Breakdown} & \it Boundary & \it Type & \it ReplyTo & \it Parent \\
\hline
Distinct users & $87$K   & Creation      & $22\%$&$100\%$&$100\%$&$100\%$ &$100\%$\\
Talk Pages     & $2.2$M  & Addition      & $50\%$&$96\%$&$100\%$&$100\%$ &$100\%$\\
Revisions      & $4.6$M  & Modification  & $9\%$&$84\%$&$94\%$&$99\%$ &$97\%$\\
Conversations  & $4.4$M  & Deletion      & $16\%$&$99\%$&$90\%$&$100\%$&$98\%$\\
Actions        & $6.4$M & Restoration   & $3\%$ &$97\%$&$98\%$&$100\%$&$98\%$\\
\cline{3-8}
 & & \multicolumn{2}{r|}{\bf All actions:}      &$96\%$&$98\%$&$99\%$&$99\%$\\ 
\hline
\end{tabular}
\caption{\label{tab-stats} \textbf{Summary statistics and reconstruction accuracy for the English and Chinese Wikipedia talk page corpora.}
These statistics exclude actions that result in empty content after markup cleaning (e.g., purely formatting edits).
}
\normalsize
\end{table*}

\section{Conversation Reconstruction}\label{sec:Dataset}

Technically, comments are added to Wikipedia talk pages the same way content is added to article pages: contributors simply edit the markdown of any part of the talk page without relying on any functionality specialized for structuring the conversations.
Figure \ref{talk-example} gives an example of the discussion interface and the resulting rendered conversation.
Each edit results in a revision of the whole page that is permanently stored in a public historical record.\footnote{In some rare cases revisions are deleted, for example, if personal information is accidentally written into a page.} 
Because conversations on Wikipedia have no `official' underlying structure, and instead are organized using indentation markup and other ad hoc visual cues, computational heuristics are necessary to interpret conversational structure.\\
\textbf{Actions.}
We model the conversational structure of interactions as a graph of {\em actions}, as illustrated in Figure~\ref{toy-example}. 
Actions are categorized into five {\em types}:\\
$\bullet$ \textit{Creation}: A conversation thread is started by adding a markup section heading (e.g., Action~1 in Figure~\ref{toy-example}).\\
$\bullet$ \textit{Addition}: A new comment is added to a thread (e.g., Actions~2 and 3).\\
$\bullet$ \textit{Modification}: An existing comment is modified (e.g., Action~5); the Parent-id indicates the original comment. \\
$\bullet$ \textit{Deletion}: A comment or thread-heading is being removed (e.g., Action~4); Parent-id specifies the comment or thread-heading's most recent action.\\
$\bullet$ \textit{Restoration}: A deletion is being reverted, returning to the state indicated by the Parent-id. \\
\noindent All action types except thread creations, thread deletions and thread restorations also include a ReplyTo-id indicating the target of the reply.\\
\textbf{From Page Revisions to Actions.}
Our reconstruction pipeline is a Python program written for Google Cloud Dataflow (also known as Apache Beam)\footnote{\url{https://cloud.google.com/dataflow/}} that operates on pages in parallel and on the revisions of each page sequentially in temporal order.

Due to the large scale of Wikipedia data, we use external sorting for pages that contains too many revisions to fit in a Dataflow worker's memory. 
When the number of revisions is too large for a Dataflow worker's local disk, the computation is performed in stages, a few years at a time.

Given the sorted set of a page-revisions, token-level diffs between sequential revisions are computed using a longest common sequence (LCS) algorithm.\footnote{\url{github.com/google/diff-match-patch}}
Each sequential diff is then decomposed into the set of atomic conversation {\em actions} attributed to the user who submitted the page revision. During the sequential processing of a page's revisions, two data structures are maintained: each comment's current character offset, and a list of deleted comments. 
The comment offsets are used to interpret the difference between modification actions (edits within the bounds of an existing action) and additions; the deleted comments are used to identify restoration of comments. 

We store the most recent 100 deleted comments between 10 to 1000 characters long, for each page. 
This is used to compute when a comment is restored by looking up deleted comments in a trie.
The token length lower bound parameter avoids short commonly added comments---like ``Thanks!''---from being interpreted as restorations.
The upper bound ensures that occasional very long deleted comments are skipped, to bound Dataflow workers' memory usage.

Finally, reconstructed actions are processed using {\em mwparserfromhell}\footnote{\url{github.com/earwig/mwparserfromhell}} to clean the MediaWiki formating.
Note that, since arbitrary page changes are allowed, some actions cannot be processed by the parser (about 1 in 200,000); in such cases, an action's raw MediaWiki markup is stored.

Table~\ref{tab-stats} shows summary statistics of the final dataset on English and Chinese Wikipedia.
The version of the raw data dumps processed were retrieved on July 1st 2018.
\section{Evaluation of Reconstruction Quality}\label{sec:Evaluation}
We evaluate the quality of the automatic reconstruction by manually verifying a randomly drawn subset of 
 (at least) 100 examples from each action category. For each action we verify the accuracy of (1) the assigned action type, (2) the token-level boundary of the comment, (3) the ReplyTo relation and (4) the action's Parent relation.

We conduct the evaluation for both English and Chinese data (Table~\ref{tab-stats}).
With over $98\%$ of actions classified correctly in both languages, the dataset exhibits a high annotation quality given its scale and detail.
From the error cases in the English data, 10\% result from limitations in the current technologies for HTML parsing and LCS matching.
User behavior that we could interpret but is not yet captured by our algorithm, such as moving ongoing conversations to another talk pages accounts for another 24\%.
The remaining errors were from edits that we were unable to interpret.  By open sourcing the reconstruction code, we encourage further refinements.
\begin{figure*}[htb!]
  \includegraphics[width=\textwidth]{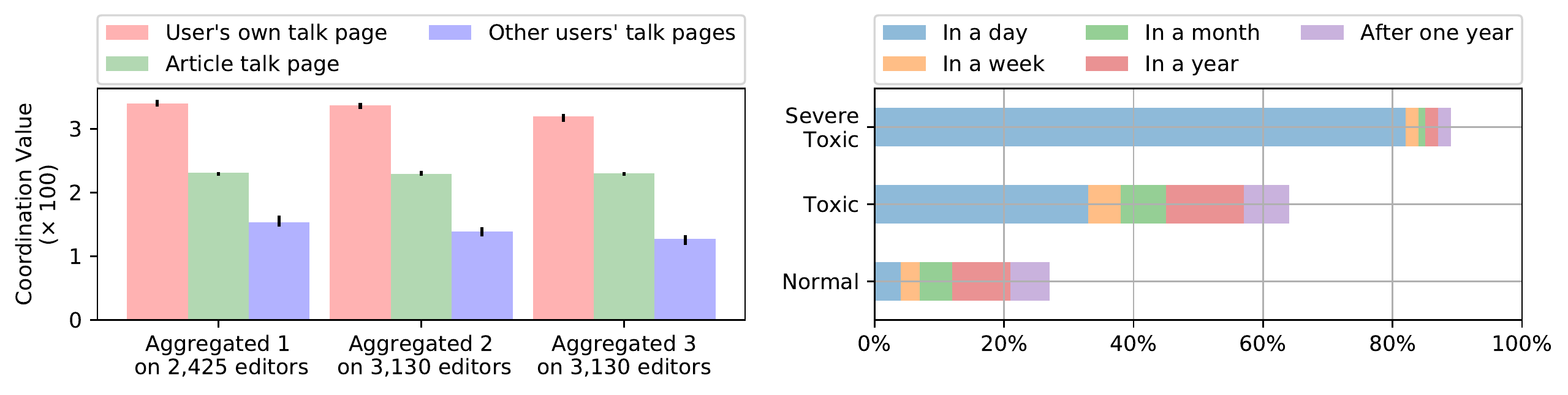}
  \caption{\label{coord}
\textbf{(Left) Linguistic coordination depends on the discussion's venue.} 
Error bars are estimated by bootstrap resampling.
\textbf{(Right) Deletion rate of content over varying time periods.}}
\end{figure*}
\section{Case Studies}\label{sec:Analysis}
We now briefly present two studies on English Wikipedia that highlight the importance of (1) collecting the full history of Wikipedia across all pages and (2) capturing the various types of interactions. 

\noindent {\bf Linguistic Coordination.}
\citet{danescu2012echoes} studied language coordinations (i.e., in a conversation between $a$ and $b$, to what degree is $b$ systematically adopting $a$'s language patterns when replying to $a$) on a conversational corpus derived from $5,657$ \textit{User Talk pages}: those associated with, and managed by, a specific user.
The study showed that social status mediates the amount of linguistic coordination, with contributors imitating more the linguistic style of those with higher status in the community.

We now show that the coordination pattern of the page owners in the previous dataset differs significantly based on where the conversation takes place.
We compare each contributor's coordination patterns on their own user talk page to patterns exhibited on talk pages of other contributors, as well as to those on article talk pages---talk pages associated with a Wikipedia article.
To avoid confounding different populations (and fall into the trap of Simpson's paradox), we only include in the comparison users that had a sufficient amount of contributions across all three venues.
Figure~\ref{coord} shows the three aggregated coordination values computed by applying the methodology of the original paper on 4 million addition actions that occurred before 2012.

Our results show with significant difference ($p<0.001$ calculated by one-way ANOVA) that contributors coordinate the least when replying on other users' talk pages, and most on their own talk page. 
This leads us to speculate a new hypothesis: contributors have a different 
perception of status or respect on their own page than on others.
Such questions, which require more thorough investigation that depends on observing how contributors interact across different discussion venues, can be studied using the WikiConv corpus.

\noindent {\bf Moderation of toxic behavior.}
\citet{wulczyn2017ex} measured prevalence of personal attacks in a Wikipedia talk page corpus, and evaluated the fraction of attacks that moderators follow up on with a block or warning ($17.9\%$).
However, because there was no structured history of comment deletion,
the authors were unable to measure the rate at which toxic comments are moderated through deletion. Using the more complete datasets provided by WikiConv we show that the fraction of problematic comments moderated by Wikipedians is significantly higher than their initial estimate suggests.

We used the Perspective API\footnote{\url{https://www.perspectiveapi.com}} to score the toxicity of all addition and creation actions (which we refer to as ``comments'' here).\footnote{We release the scores with the dataset.}
Each comment is further classified as toxic or non-toxic according to the equal error rate threshold, following the methodology of~\cite{wulczyn2017ex}, where false positives are offset by false negatives. 
The threshold is calculated by on the human labels in the Kaggle Toxicity  dataset of Wikipedia comments.\footnote{The Jigsaw Toxicity Kaggle Competition: \url{goo.gl/N6UGPK}} 
Classification at this threshold yields $86\%$ precision and $84\%$ recall.

We used the same method to labeled comments with the \emph{severe toxic} model.
Figure \ref{coord} shows the fraction of comments deleted by Wikipedians who are not the author of the comment for different lengths of time; distinguishing between comments labeled as toxic, severely toxic, and the background distribution.
The key observations here are that nearly $33\%$ of toxic comments are removed within a day; and over 82\% of severely toxic comments are deleted within a day.
This complements results previously reported by \citet{wulczyn2017ex}, accounting for an additional type of community moderation that is revealed using the detailed information about the history of the conversation provided by our corpus.

\section{Conclusion and Future Work}

We introduced a pipeline that extracts the complete conversational history of Wikipedia talk pages at a level of detail that was not previously available. 
We applied this pipeline to Wikipedia in multiple languages and evaluated its quality on the English and Chinese Talk page corpora, 
obtaining
a high reconstruction accuracy for both the Chinese and English datasets (98\%).
This level of detail and completeness opens avenues for new research, as well as for revisiting and extending existing work on online conversational and collaboration behavior. 
For example, while in our use cases we have focused on contributors deleting toxic comments, one could seek to understand why and when an editor is deleting or rewording their own comments. 
Beyond refining the heuristics and parsing methods used in our reconstruction pipeline, and reducing the time to update the corpus, a remaining challenge is to capture conversations that happen across page boundaries.
\section{Acknowledgement}

We thank Thomas Ristenpart, Andreas Veit for proof reading; Ben Vitale for many helpful discussions on building the pipeline; and Jonathan P. Chang for reporting data issues and discussing the challenges throughout.
This project is supported in part by NSF grant CNS-1558500.
\bibliography{emnlp2018}
\bibliographystyle{acl_natbib_nourl}
\end{document}